\documentclass{llncs}

\usepackage[utf8]{inputenc} % allow utf-8 input
\usepackage[T1]{fontenc}    % use 8-bit T1 fonts
\usepackage{hyperref}       % hyperlinks
\usepackage{url}            % simple URL typesetting
\usepackage{booktabs}       % professional-quality tables
\usepackage{amsfonts}       % blackboard math symbols
\usepackage{amsmath}
\usepackage{amssymb} 
\usepackage{nicefrac}       % compact symbols for 1/2, etc.
\usepackage{microtype}      % microtypography
\usepackage{lipsum}
\usepackage{xcolor}
\usepackage{graphicx}
\usepackage[ruled,vlined,linesnumbered]{algorithm2e}
\usepackage[caption=false,font=footnotesize]{subfig}
\usepackage{changepage}

\newcommand{\knn}{$k$-NN}

\newcommand{\argmax}{\mathop{\arg\max}}

\let\emptyset\varnothing

\title{Cluster Representatives Selection in Non-Metric Spaces for Nearest Prototype Classification}
\titlerunning{Cluster Representatives Selection}

\begin{document}
\author{Jaroslav Hlav{\'a}{\v c}\inst{1,2} \and Martin Kopp\inst{1,2} \and Jan Kohout\inst{1,3}}
%\author{For review}

\institute{Cognitive Intelligence, Cisco Systems, Prague, Czech Republic. \and
Faculty of Information Technology, Czech Technical University in Prague, Czech Republic. \and
Faculty of Electrical Engineering, Czech Technical University in Prague, Czech Republic.}

\maketitle

\begin{abstract}
The nearest prototype classification is a less computationally intensive replacement for the \knn{} method, especially when large datasets are considered. 
In metric spaces, centroids are often used as prototypes to represent whole clusters.
The selection of cluster prototypes in non-metric spaces is more challenging as the idea of computing centroids is not directly applicable.

In this paper, we present CRS, a novel method for selecting a small yet representative subset of objects as a cluster prototype.
Memory and computationally efficient selection of representatives is enabled by leveraging the similarity graph representation of each cluster created by the NN-Descent algorithm.
CRS can be used in an arbitrary metric or non-metric space  because of the graph-based approach, which requires only a pairwise similarity measure.
As we demonstrate in the experimental evaluation, our method outperforms the state of the art techniques on multiple datasets from different domains.
\end{abstract}

% keywords can be removed

\section{Introduction}
\label{sec:intro}
The \knn{} classifiers are often used in many application domains due to their simplicity and ability to trace the classification decision to a specific set of samples. 
However, their adoption is limited by high computational complexity. 
Because contemporary datasets are often huge, containing hundreds of thousands or even millions of samples, computing similarity between the classified sample and the entire dataset may be computationally intractable.

In order to decrease computational and memory requirements, the nearest prototype classification (NPC) method is commonly used, c.f.~\cite{seo2003soft,schleif2005local,cervantes2007adaptive}.
In NPC, each class is divided into one or more clusters, and each cluster is represented by its \emph{prototype}.
The classified sample is then compared just to the prototypes instead of calculating similarity to the entire dataset. 

Therefore, the goal of prototype selection is to find a memory-efficient representation of clusters such that classification accuracy is preserved while the number of comparisons is significantly reduced. 

However, in many application domains, objects might exist in a non-metric space where only a pairwise similarity is defined, e.g., bioinformatics~\cite{martino2018granular}, biometric identification~\cite{becker2010methods}, computer networks~\cite{kopp2018community} or pattern recognition~\cite{scheirer2014good}.

In such application domains, standard representations such as centroids may not be easily determined, or their interpretation does not make much sense. 
For these scenarios, only a few methods have been developed, and to best of our knowledge the only general (not domain-specific) approach is based on the selection of small subsets of objects to represent the remaining cluster members. These object, called \emph{representatives}, are then used as a prototype.

While several methods capable of solving representatives selection on non-metric spaces exist (i.e. DS3~\cite{elhamifar2015dissimilarity}, $\delta$-medoids~\cite{liebman2015representative}), there has not been much research activity in this direction.

Our focus on non-metric spaces comes from the problem of behavioural clustering of network hosts~\cite{kopp2018community}. 
Nevertheless, the problem of selecting a minimal number of representative samples is of more general interest. 
Therefore, we present a novel method to solve the problem of Cluster Representatives Selection (CRS).
CRS is a general method capable of selecting small representative subset of objects from a cluster as its prototype.
Its core idea is fast construction of an approximate reverse \knn{} graph and then solving minimal vertex cover problem on that graph.
Only a pairwise similarity is required to build the reverse \knn{} graph, therefore application of CRS is not limited to metric spaces.

To show that CRS is general and domain-independent, we present an experimental evaluation on datasets from image recognition, document classification and network host classification, with appealing results when compared to the current state of the art. 

The paper is organized as follows. The related work is briefly reviewed in the next section. Section~\ref{sec:problem} formalises the representative selection as an optimization problem. The proposed CRS method is described in detail in Section~\ref{sec:crs}. The experimental evaluation is summarized in Section~\ref{sec:experiments} followed by the conclusion.

\section{Related Work}
\label{sec:related}
During the past years, significant effort has been made to represent clusters in the most condensed way. The approaches could be categorized into two main groups.

The first group gathers all prototype generation methods~\cite{triguero2011taxonomy}, which create artificial samples to represent original clusters, e.g.~\cite{geva1991adaptive,xie1993vector}. The second group contains the prototype selection methods. As the name suggests, a subset of samples from the given cluster is selected to represent it. Prototype selection is a well-explored field with many approaches, see, e.g.~\cite{garcia2012prototype}.

However, most of the current algorithms exploit the properties of the metric space, e.g., structured sparsity~\cite{wang2017representative}, $l_1$-norm induced selection~\cite{zhang2018seeing} or identification of borderline objects~\cite{olvera2018accurate}.

When we leave the luxury of the metric space and focus on situations where only a pairwise similarity exists or where averaging of existing samples may create an object without meaning, there is not much previous work.

The $\delta$-medoids~\cite{liebman2015representative} algorithm uses the idea of $k$-medoids to semi-greedily cover the space with $\delta$-neighbourhoods, in which it then looks for an optimal medoid to represent a given neighbourhood.
The main issue of this method is the selection of $\delta$: this hyperparameter has to to be fine-tuned based on the domain.

The DS3~\cite{elhamifar2015dissimilarity} algorithm calculates the full similarity matrix and then selects representatives by a row-sparsity regularized trace minimization program which tries to minimize the rows needed to encode the whole matrix. The overall computational complexity is the most significant disadvantage of this algorithm, despite some proposed approximate estimation of the similarity matrix using only a subset of the data. 

The proposed method for Cluster Representatives Selection (CRS) approximates the topological structure of the data by creating a reverse \knn{} graph.
CPS then iteratively selects nodes with the biggest reverse neighbourhoods as representatives of the data.
This approach systematically minimizes the number of pairwise comparisons to reduce computational complexity while accurately representing the data.

\section{Problem Formulation}
\label{sec:problem}
In this section, we define the problem of prototype-based representation of clusters and the nearest prototype classification (NPC). As we already stated in Introduction, we study the prototypes selection in general cases, including non-metric spaces. Therefore, we further assume that a cluster prototype is always specified as (possibly small) subset of its members.

\paragraph{Cluster prototypes}
Let $\mathrm{T}$ be an arbitrary space of objects for which a pairwise similarity function $s: \mathrm{T} \times \mathrm{T} \to \mathbb{R}$ is defined and let $X \subseteq \mathrm{T}$ be a set of (training) samples. Let $\mathcal{C} = \{C_1, ..., C_m\}$ be a clustering of $X$ such that $C_i \cap C_j = \emptyset, \forall i\neq j$ and $\bigsqcup C_i = X.$
Let $C_i = \{x_1, x_2, ..., x_n\}$ be a cluster of size $n$.
For $x \in C_i$, let us denote $U^k_x$ the $k$ closest samples to $x$, i.e., the set of $k$ samples that have the highest similarity to $x$ in the rest of the cluster $C_i \setminus \{ x \}$.
Then the goal of the prototype selection is to find a subset of samples $R_i \subseteq C_i$ for each cluster such that:
\begin{equation}
\label{eq:coverage}
\forall x \in C_i \; \exists \; r \in R_i \;:\; x \in U^k_r 
\end{equation}
The set $R_i$ is then called the prototype of the cluster $C_i$. In case of ties, we pick the samples with lowest indices $i$.

In order to minimize computational requirements of NPC, we search for a minimal set of cluster representatives $R_i$ for each cluster, which satisfies the coverage requirement~\eqref{eq:coverage}:
\begin{equation}
\label{eq:optim}
% WIP (older versions of the eqn below)
%R_i^* = \argmin_{|R_i|} \;  \left\{ \bigcup_{r \in R_i} U^k_r = C_i \right\}
%\left| R_i^* \right| = \min_{|R_i|} \;  \left\{ \bigcup_{r \in R_i} U^k_r = C_i \right\}
\left| R_i \right| = \min_{R} \;  \left|  \left\{ \bigcup_{r \in R} U^k_r = C_i \right\} \right|
\end{equation}
Note that several sets might satisfy this coverage requirement.

\paragraph{Relaxed prototypes}
Finding cluster prototypes that fully meet the coverage requirement~\eqref{eq:coverage} might pose an unnecessary computational burden.
In many cases, a smaller prototype which is much easier to obtain can capture enough of the important characteristics of a cluster despite possibly not covering all of its members (e.g., a few outliers). Motivated by this observation, we introduce a relaxed requirement on cluster prototypes.
We say that a set $R_i \subseteq C_i$ is a representative prototype of cluster $C_i$ if the following condition is met:
\begin{equation}
\label{eq:relaxed_coverage}
\left|\bigcup_{r \in R_i}U_r^k \cap C_i \right| \geq \epsilon|C_i|,
\end{equation}
for a preset parameter $\epsilon \in (0,1]$.
%\jknote{Alternatively, we could also consider a probabilistic formulation $$\forall x \in C_i: \; P\left(x \in \bigcup_{r \in R_i}U_r^k\right) \geq \epsilon
%$$}

In further work, we replace the requirement~\eqref{eq:coverage} with its relaxed version~\eqref{eq:relaxed_coverage}. In case of need, the full coverage requirement can be enforced by simply setting $\epsilon = 1$. Similarly, also in the relaxed version, we seek for a prototype with minimal cardinality which satisfies~\eqref{eq:relaxed_coverage}.

\paragraph{Nearest Prototype Classification}
Having the representative prototypes for all clusters, we now describe how the classification is performed.
In the nearest prototype classification (NPC), an unseen sample $x$ is classified to the cluster (i.e., the respective target class is assigned to the sample) which is represented by the prototype with the highest similarity to the sample $x$.
As the cluster prototypes are disjoint sets, the nearest prototype is defined as the prototype containing the sample with the highest similarity to $x$. Formally, given the prototypes of all clusters $\mathcal{R} = \{R_1,...,R_m\}$, the nearest prototype, denoted $R^{*}$, is the prototype containing  $r^{*}$, where
$$
r^{*} = \argmax_{r \in \bigcup R_i} s(x,r).
$$

Again, we resolve ties by picking the candidate with lowest index in the dataset.

Finally, the sample $x$ is classified with the same label as that of the cluster represented by the prototype $R^{*}$.

%---------------------------------
%classification searches for the most similar $r_{sim}$ sample in all prototypes~\eqref{eq:npc}:
%\begin{equation}
%\label{eq:npc}
%r_{sim} = \argmax_{r \in \cup_{r \in R_i}}\{\textrm{sim}(t, r)\},
%\end{equation} 
%where $\textrm{sim}$ is the pairwise similarity used.
%Sample $t$ is then assigned to a cluster $R_{sim}$ which $r_{sim}$ is a member of.

 %More formally, given the set of prototypes of all clusters $\mathcal{R} \in \{R_1,…,R_m\}$, the nearest prototype for sample $x$, denoted $R^{x}$, is the prototype which contains the nearest representative $r^{x}$, where
%\begin{equation}
%\label{eq:npc2}
%r^{x} = \argmax_{r \in \cup R_i} s(x,r),
%\end{equation}

\section{Cluster Representatives Selection}
\label{sec:crs}

\begin{figure}[ht]
\centering
\subfloat[dataset\label{fig:data}]{%
  \includegraphics[width=0.33\textwidth]{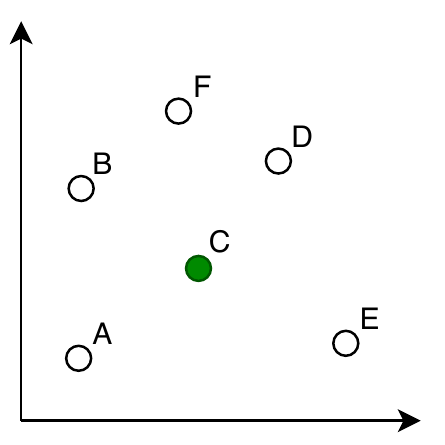}
  }%
\subfloat[k-neighbourhood\label{fig:neighbourhood}]{%
  \includegraphics[width=0.30\textwidth]{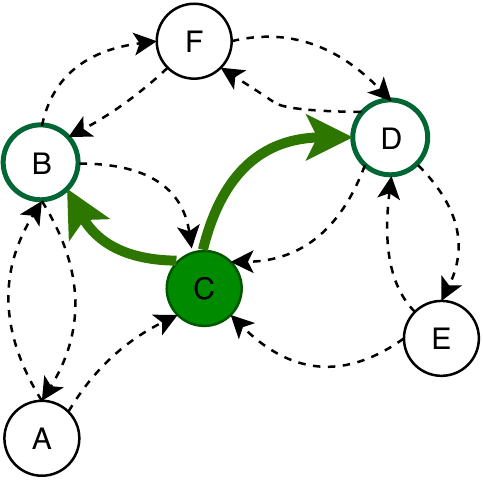}
}%
\subfloat[reverse neighbourhood\label{fig:reverse}]{
  \includegraphics[width=0.30\textwidth]{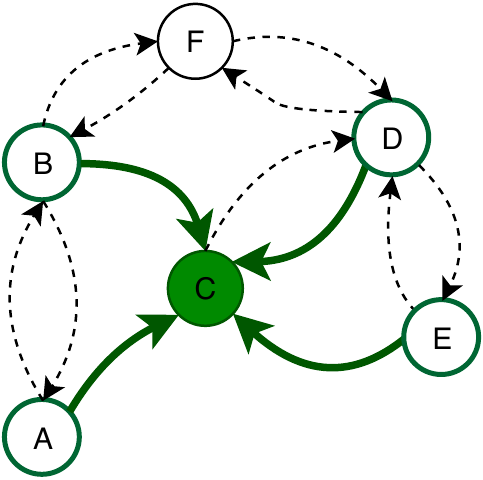}
}%
\caption{Illustration of the steps of CRS algorithm. (a) Visualization of a toy 2D dataset. (b) 2-NN graph created from from the dataset. (c) Reverse graph created from the graph depicted in (b). Point C is a representative of A, B, D, E and would be a good choice first choice of a representative. Depending on the coverage parameter $\epsilon$, the node F could be considered an outlier or also added to the representation.}
\label{fig:crs}
\end{figure}

In this section, we describe our method CRS for building the cluster prototypes. The entire method is composed of two steps that are discussed in more detail in individual subsections.
First, given a cluster $C$ and a similarity measure $s$, a  reverse \knn{} graph  $G$ is constructed from objects  $C$ using the pairwise similarity $s$.
Then, the graph $G$ is used to select the representatives that satisfy the coverage requirement while minimizing the size of the cluster prototype.
The simplified scheme of the whole process is depicted in Figure~\ref{fig:crs}.

\subsection{Building the Prototype}
\label{sec:prototype}
For the purpose of building the prototype for a cluster $C$ a weighted reverse \knn{} graph $G^{-1}_C$ is used. 
It is defined as $G^{-1}_C = (V, E, w)$, where $V$ is the set of all objects in cluster $C$, $E$ is a set of edges and $w$ is a weight vector. 
An edge between two nodes $v_i, v_j \in V_{i \neq j}$ exists if $v_i \in U_{v_j}^k$, while the edge weight $w_{ij}$ is given by the similarity $s$ between the connected nodes, $w_{ij} = s(v_i, v_j)$.

The effective construction of such graph is enabled by employing the NN-Descent~\cite{dong2011efficient} algorithm which produces a \knn{} graph $G_C$. The reverse \knn{} graph $G^{-1}_C$ is then obtained from $G_C$ by simply reversing directions of the edges in $G_C$.

NN-Descent is a fast converging approximate method for the \knn{} graph construction.
It exploits the idea that ``a neighbour of a neighbour is also likely to be a neighbour'' to locally explore neighbouring nodes for better solutions.

Having the reverse \knn{} graph $G_C^{-1}$, we want to ensure that each object $x$ is at least $\tau$-similar to all its neighbours, i.e.,
\begin{equation*}
\left(\forall y \in U_x : s(x,y) \geq \tau \right).
\end{equation*}
Omitting all edges with weight lower than $\tau$ not only lowers the memory requirements, but it also unfolds objects with large neighbourhood as good representative candidates.

The selection of representatives is treated as a minimum vertex cover problem on $G_C^{-1}$.
We use a greedy algorithm which iteratively selects objects with maximal $|U|$ as representatives and marks them and their neighbourhood as covered.
The algorithm stops when the coverage requirement \eqref{eq:relaxed_coverage} is met (see Section~\ref{sec:problem}).

The whole algorithm is summarized in Algorithm~\ref{alg:crs}.

\begin{algorithm}[ht]
\SetAlgoLined
\KwData{cluster $C=\{c_1, ..., c_n\}$, similarity $s$, coverage threshold $\epsilon$}
\KwResult{set of selected representatives $R \subseteq C$}
$G_C =$ NN-Descent($C$, $s$) \\
$G_{C}^{-1} = $ ReverseGraph($G_C$) \\
$Z = C$ \quad //\textit{set of uncovered objects} \\
$R = \{\}$ \quad //\textit{set of representatives}\\
 \While{$\frac{|C|-|Z|}{|C|} < \epsilon$}{
   $r = \argmax(\sum\limits_{c \in Z} s(c, u), u \in U_c$) \\
   $Z = Z \setminus U_{r}$ \\
   $R = R \cup \{r\}$ \\
 }
 \textbf{return} $R$ 
 \caption{Pseudocode for Cluster Representatives Selection}
 \label{alg:crs}
\end{algorithm}

\begin{figure}
    \centering
    \includegraphics[width=\textwidth]{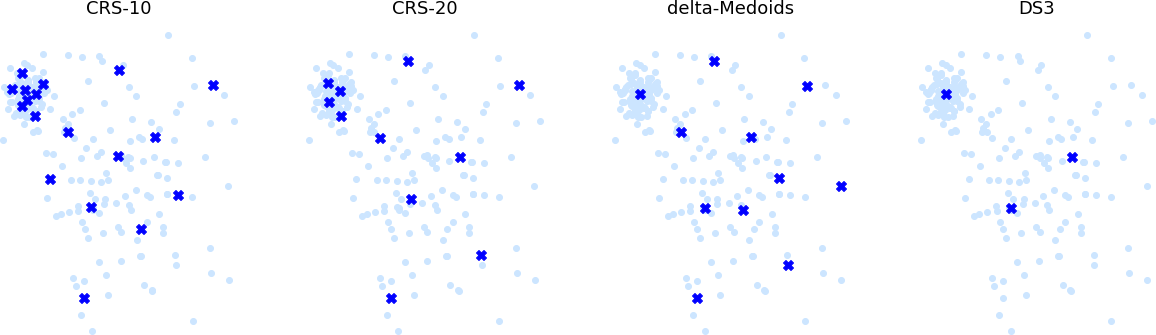}
    \caption{Selection of representatives CRS with different $K$s for a 2-dimensional dataset with 255 samples. For better comparison $\delta$-medoids and DS3 are also shown. CRS takes into account the density of different parts of the cluster and selects representatives accordingly. $\delta$-Medoids covers the dataset entirely by evenly distributed representatives. DS3 in selects the least representatives but does not capture the overall structure of the cluster very well.}
    \label{fig:struct}
\end{figure}

An example of a cluster prototype selected by the CRS algorithm is presented in Figure~\ref{fig:struct}.

\subsection{Discussion on parameters}
\label{sec:params}
This subsection summarizes the parameters of the CRS method.
\begin{itemize}
\item $k$: number of neighbors for the \knn{} graph creation. When $k$ is high, each object covers more neighbours, but on the other hand it also increases the number of pairwise similarity calculations. This trade-off is illustrated for different values of $k$ in Figure~\ref{fig:spirals}.
Due to the large impact of this parameter on properties of the produced representations and computational requirements, we further study its behaviour in more detail in a dedicated experiment in Section~\ref{sec:experiments}. 

\begin{figure}
    \centering
    \includegraphics[width=\textwidth]{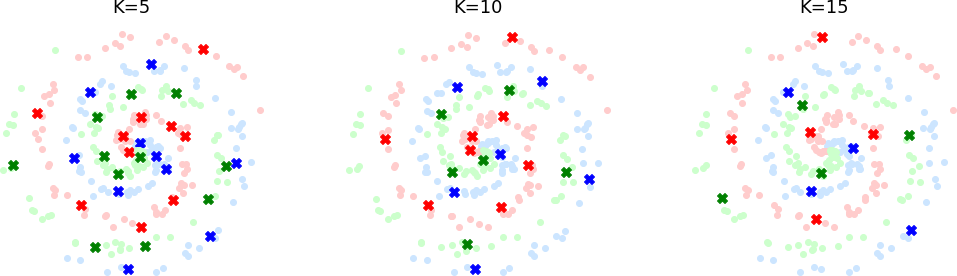}
    \caption{Number of representatives being selected and the quality of representation are both controlled by $k$. As each object explores a bigger neighbourhood for higher $k$, the number of other objects it represents grows, therefore the number of representatives decreases. On the other hand, with less representatives, some information about the structure is lost, as in the case of $k=15$.}
    \label{fig:spirals}
\end{figure}

\item $\epsilon$: coverage parameter for the relaxed coverage requirement as introduced in Section~\ref{sec:problem}. In this work, we set it to 0.95 such that the vast majority of each cluster is still covered but outliers do not influence the prototypes.
\item $\tau$: threshold on weights, determining which edges will be kept in the graph $G_C^{-1}$ (see Section~\ref{sec:prototype}). By default it is automatically set to the value of \emph{homogeneity} $h(C)$ of the cluster $C$:
\begin{equation}
\label{eq:homogeneity}
h(C) = {1 \over |C|(|C|-1)}\sum_{x_i,x_j \in C, i \neq j}s(x_i,x_j)
\end{equation}
\end{itemize}

Additionally, the NN-Descent algorithm, used within the CRS method, has two more parameters that specify its behaviour during the \knn{} graph creation. First, the $\delta_{nn}$ parameter which is used for early termination of the NN-Descent algorithm when the number of changes in the constructed graph is minimal. We set it to 0.001, as suggested by the authors of the original work~\cite{dong2011efficient}.
Second, the sample rate $\rho$ controls the number of reverse neighbours to be explored in each iteration of NN-Descent. Again, we set it to 0.7 to speed up the \knn{} creation while not diverging too far from the optimal solution in accordance with suggestions published in~\cite{dong2011efficient}.

%While tuning parameters for UPS we are trying to balance the trade-off in the number of pairwise similarities that need to be computed and ability to capture the correct structure of the data.
%Both of these properties are contained in the value of $K$.

%The maximum of comparisons we want to make in creating the graph representation of the problem is determined by the calculation of full similarity matrix.
%The upper bound on the number of comparisons of the NN-descent algorithm is set by:
%\begin{enumerate}
%    \item $\log_2(n)$ - the number of cycles the algorithm performs
%    \item $k^2$ - the upper bound of number of comparisons made for each node each cycle
%    \item $n$ - the number of nodes
%\end{enumerate}.

%Thus we create the following inequality:
%\begin{equation}
%     k^2 n \log_{2}(n)  \leq  \frac{n(n-1)}{2}
%    \label{eq:K_solve}
%\end{equation}
%By solving this inequality we can estimate $k$ as:
%\begin{equation}
%    k \leq \sqrt{ \frac{n-1}{2 \log_{2}(n)}}
%    \label{eq:K}
%\end{equation}

%We set a lower bound on $k$ as 5.
%Creating a $k$-NN graph for lower $k$s does not make sense as each sample would not get to explore any neighborhood.

\section{Experiments}
\label{sec:experiments}

This section presents experimental evaluation of the CRS algorithm on multiple datasets from very different domains that cover computer networks, text documents processing and image classification.
In the first  experiment, we study the  influence of  the  parameter $k$ (which  determines  the number of nearest neighbors used for building the underlying \knn{}  graph). Next, we  compare the CRS  method to the state of the art techniques DS3\cite{elhamifar2015dissimilarity} and $\delta$-medoids~\cite{liebman2015representative} in the nearest prototype  classification task on different datasets.

We set $h$ as an approximate homogeneity calculated from random 5\% of the cluster. We use $h$ as $\delta$ for $\delta$-Medoids algorithm. It makes the most sense in comparing with CRS, because CRS is also restricting the similarity by $h$ in reverse graph creation. The best results for DS3 we obtained with $p=\inf$ and $\alpha=3$, while creating the full similarity matrix for the entire cluster. Finally the parameters for CRS are discussed in Section~\ref{sec:params}. The following experiment explores the impact of $k$ in greater detail.

\subsection{Impact of $k$}
\label{sec:impact-k}
\begin{figure}[h]
\centering
\subfloat[MNIST Fashion - Dress\label{fig:mnist-c1}]{%
  \includegraphics[width=0.5\textwidth]{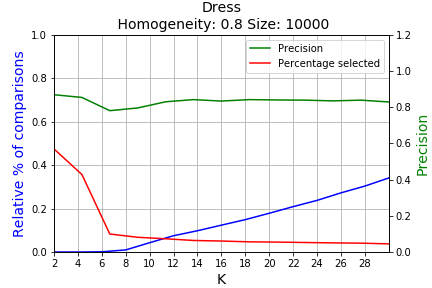}
  }%
  \subfloat[Medium Network Cluster\label{fig:network-c1}]{
  \includegraphics[width=0.5\textwidth]{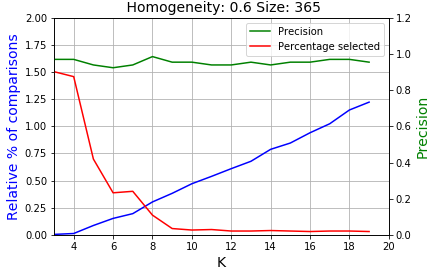}
}%
\\
\subfloat[MNIST Fashion - Sandal\label{fig:mnist-c2}]{%
  \includegraphics[width=0.5\textwidth]{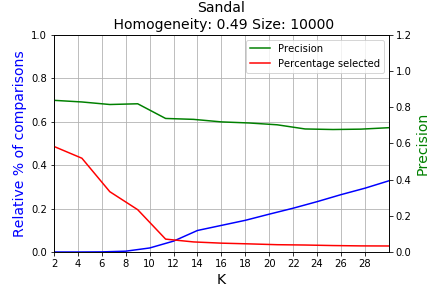}
}%
\subfloat[Big Network Cluster\label{fig:network-c2}]{%
  \includegraphics[width=0.5\textwidth]{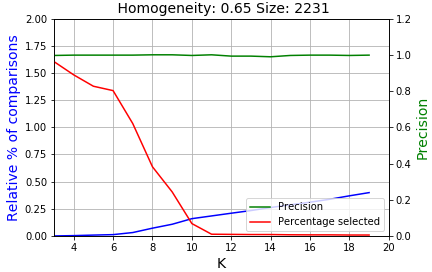}
  }%
\caption{Illustration of how the selection of $k$ influences the number of representatives and number of similarity computations. The number of representatives is in relative numbers to the size of the cluster. For different clusters as $k$ increases the relative number of comparisons also increases. However, the size of prototype selected decreases steeply while the precision only decreases slowly.}
\label{fig:setk}
\end{figure}

When building cluster prototypes by the CRS  method, the number of nearest neighbors considered for building the \knn{} graph (specified by the parameter $k$) plays very important role. With small values of $k$, each object represents only few of its neighbors that are most similar to it. However, this also increases the number of representatives needed to sufficiently cover the cluster. On the other hand, higher values of $k$ produce smaller prototypes as each representative is able to cover more objects. Nonetheless, this is at the cost of increased computational burden because the cost of \knn{} creation increases rapidly with higher $k$s. These trends can be well observed in Figure~\ref{fig:setk} which shows classification precision, sizes of created prototypes and numbers of similarity function evaluations depending on $k$ for several clusters that differ in their homogeneity and sizes. We can see the changing trade-off between computational requirements (blue line) and memory requirements (red line) as the $k$ increases. However, this is mostly without significant impact on classification precision. The parameter $k$ can be therefore set depending on the preferences on computational requirements without significantly decreasing the classification performance.

\subsection{Datasets}
In this section we briefly describe the three datasets used in the ongoing subsections for experimental comparison of individual methods.

\subsubsection{MNIST Fashion}
The MNIST Fashion~\cite{xiao2017/online} is a well established dataset for image recognition consisting of 60000 black and white images of fashion items belonging to 10 classes. It recently replaced the overused handwritten digits datasets in many benchmarks. In case of this dataset, the cosine similarity was used as the similarity function $s$.

\subsubsection{20 Newsgroup}
This dataset is a known benchmark dataset for text documents processing.
It is composed of nearly 20 thousand newspaper documents from 20 different classes (topics).
The dataset was preprocessed such that each document is represented by a TF-IDF frequency vector. As a similarity function, we again use the cosine similarity which is a common choice in the domain of text documents processing. 

\subsubsection{Private Network Dataset}
This dataset was collected on a corporate computer network, originally for the purpose of network host clustering based on their behavior~\cite{kopp2018community}. The work defines a specific similarity measure on top of network hosts which we adopt for this paper. Clusters of network hosts were defined according to results achieved in the original work as well. Additionally, for the purposes of the evaluation, clusters smaller than 10 members were not considered, since such small clusters can be easily represented by any method.
In contrast to the previous datasets, the sizes and values of homogeneity of clusters in the Network dataset differ significantly.

\subsection{Evaluation of Results}

\begin{figure}[ht]
\centering
\subfloat[MNIST precision\label{fig:mnist-prec}]{%
  \includegraphics[width=0.32\textwidth]{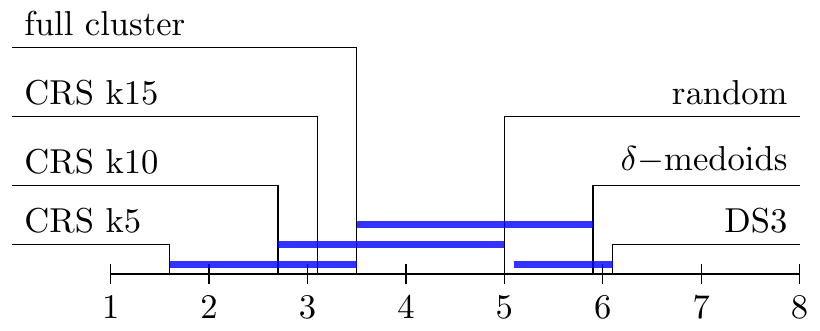}
  }%
\subfloat[news precision\label{fig:news-prec}]{%
  \includegraphics[width=0.32\textwidth]{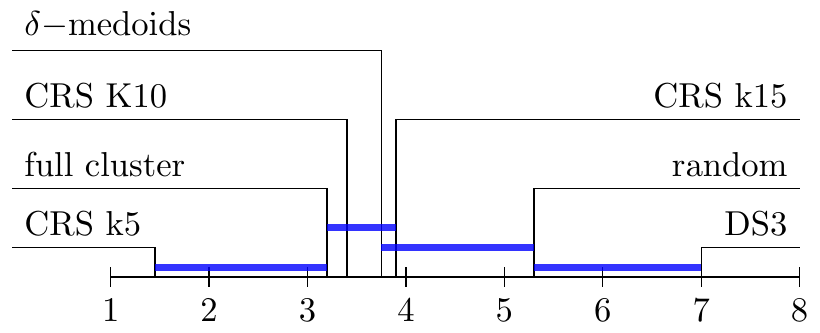}
}%
\subfloat[network precision\label{fig:network-prec}]{
  \includegraphics[width=0.32\textwidth]{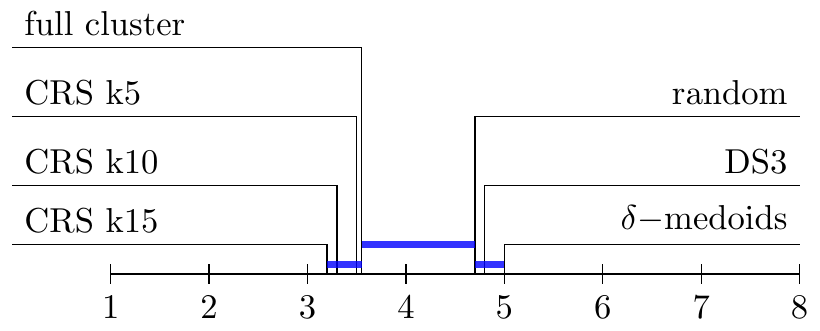}
}%
\\
\subfloat[MNIST recall\label{fig:mnist-rec}]{%
  \includegraphics[width=0.32\textwidth]{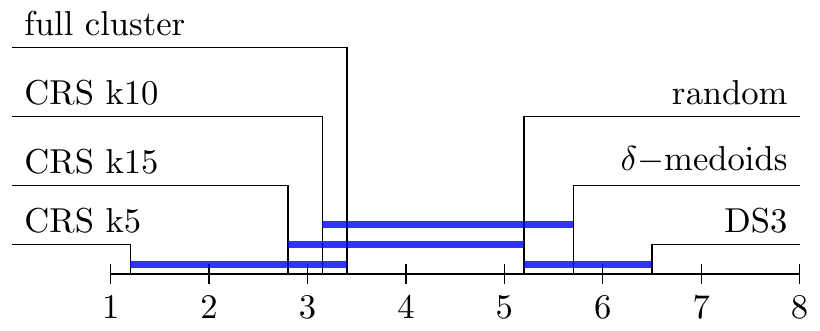}
  }%
\subfloat[news recall\label{fig:news-rec}]{%
  \includegraphics[width=0.32\textwidth]{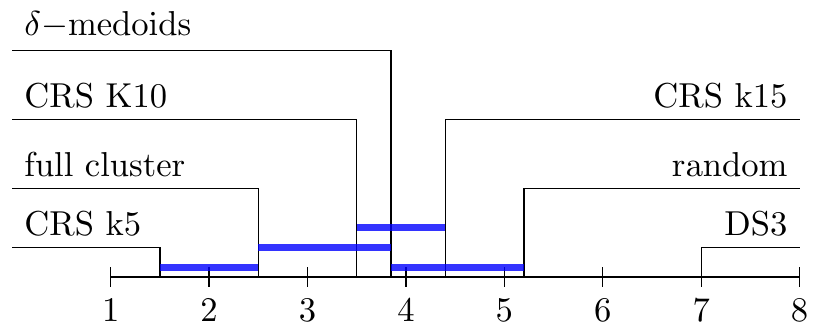}
}%
\subfloat[network recall\label{fig:network-rec}]{
  \includegraphics[width=0.32\textwidth]{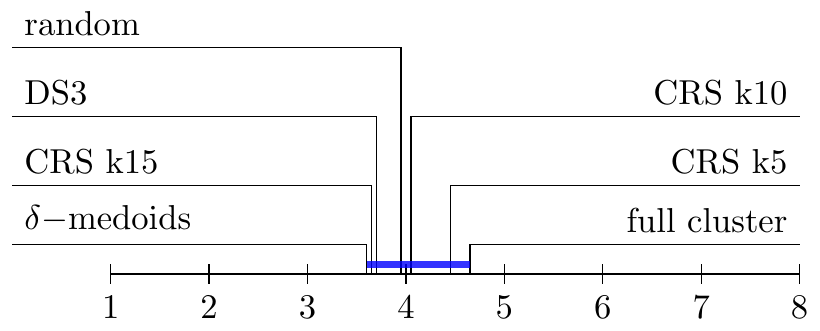}
}%
\caption{Critical difference diagram comparison (cf.~\cite{demvsar2006statistical}) of algorithms constructed using Friedman's test with correction for multiple post-hoc hypotheses by Shaffer~\cite{shaffer1995multiple}.
The diagrams show the average rank of each algorithm over all clusters in each dataset, groups of algorithms that are not significantly different (p=0.05) are connected.}
\label{fig:CDs}
\end{figure}

In this section we present the results for each dataset in detail.
The main results are summarized in Table~\ref{tab:all_res}.
For a more complete picture we also included results for a random 5\% and all 100\% of the cluster as a prototype.
The statistical comparison of the methods can be found in Figure~\ref{fig:CDs}.
Better rankings for some of CRS methods reflect, that CRS only covers $\epsilon$ which removes the outliers that decrease precision and recall of full cluster representation.

\begin{table}[]
\begin{centering}
   \begin{tabular}{|l|l|l|l|}
    \hline
    \textbf{Method}  & \textbf{MNIST Fashion} & \textbf{20Newsgroup}  & \textbf{Network}      \\ \hline \hline
    $\delta$-medoids & 0.763/0.744 (4.73\%)   & 0.542/0.515 (14.51\%) & 0.865/0.978 (7.38\%)  \\ \hline
    DS3              & 0.657/0.563 (0.07\%)   & 0.133/0.132 (0.64\%)  & 0.87/0.977 (1.88\%)       \\ \hline
    random-5\%    & 0.793/0.784 (5.0\%)  & 0.452/0.435 (5.06\%) & 0.958/0.97 (5.09\%)\\ \hline
    full-100\%    & 0.823/0.817 (100.0\%) & 0.56/0.548 (100.0\%) & 0.987/0.963 (100.0\%) \\ \hline \hline
    CRS-k5        &  0.855/0.852 (87.71\%) & 0.635/0.632 (56.58\%)   & 0.988/0.958 (65.31\%) \\ \hline
    CRS-k10       & 0.836/0.826 (15.37\%)  & 0.538/0.516 (7.14\%)  & 0.985/0.965 (7.74\%)  \\ \hline
    CRS-k15       & 0.828/0.823 (5.08\%)  & 0.522/0.488 (5.17\%) & 0.983/0.982 (5.26\%)  \\ \hline
    \end{tabular}
    \caption{Average precision/recall values for each method used on each dataset. The table also shows the percentage of the cluster that was selected as a prototype. Our algorithm is on par with existing methods while selecting noticeably fewer representatives.}
    \label{tab:all_res}
\end{centering}
\end{table}

When evaluating the experiments, we take into account both precision/recall and the percentage of samples selected as prototypes.
As we have shown in the experiment in Section~\ref{sec:impact-k}, CRS can be tuned by the parameter $k$ to significantly reduce the number of representatives and maintain a high precision/recall values.
When using the full cluster as its prototype, the average values of precision and recall are slightly lower than when using the CRS method.
This shows that CRS with $\epsilon = 0.95$ makes the classification immune to outliers which can otherwise decrease the classification quality.
The DS3 method selects a significantly lower number of representatives than any other method.
However, it is at the cost of lower precision and recall values.

Runtimes of individual algorithms also differ significantly. We evaluate the runtime requirements of each algorithm by the relative number of similarity computations $S$ defined as:
\begin{equation}
S = \frac{S_{actual}}{S_{full}},
\end{equation}
where $S_{actual}$ stands for the actual number of comparisons made and $S_{full}$ is the hypothetical number needed for computing full similarity matrix.

We use DS3 with the full similarity matrix to get most accurate results, therefore $S_{DS3} = 1$.
For $\delta$-Medoids the number of computations performed can not be easily preset.
Therefore, we averaged the number of comparisons over different values of $S_{\delta} \in [0.45, 0.7]$ (the number might differ significantly for different values of $S_{\delta}$).
For CRS the number of comparisons is influenced by $k$, homogeneity of each cluster and its size.
The impact of $k$ was discussed in detail in Section~\ref{sec:impact-k}.
The experiment shows that one can make assumptions about $S$ based on the size of the cluster and $k$ (i.e. for MNIST Fashion dataset $S_{CRS-k10} = 0.063$, $S_{CRS-k15} = 0.13$).

\subsubsection{MNIST Fashion}
\begin{figure}
    \centering
    \includegraphics[width=\textwidth]{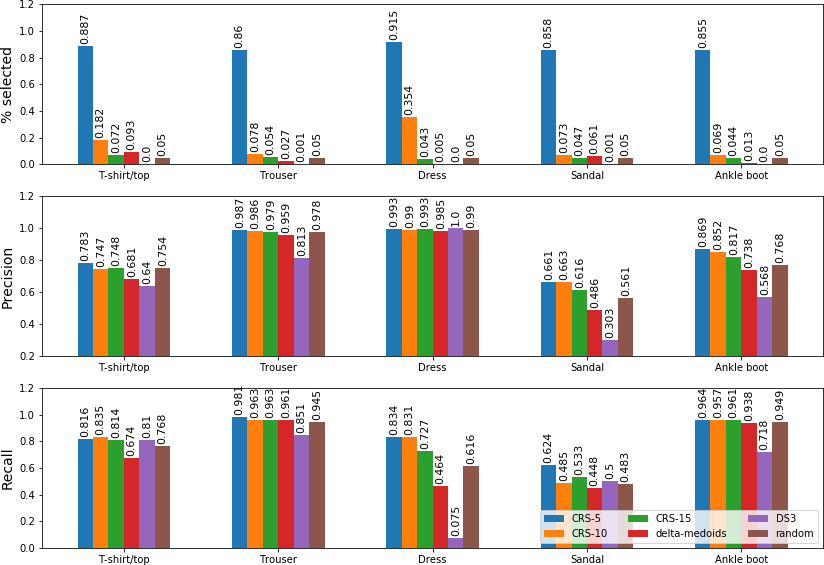}
    \caption{Visualization of precision and recall of all methods in relation to percentage of cluster selected on chosen clusters for the MNIST Fashion dataset. Values 0.0 for some clusters for DS3 mean that less than 0.1\% objects were selected as representatives.}
    \label{fig:bar-fash}
\end{figure}

\begin{figure}
\centering
\subfloat[CRS-k10\label{fig:fash-crs}]{%
  \includegraphics[width=0.32\textwidth]{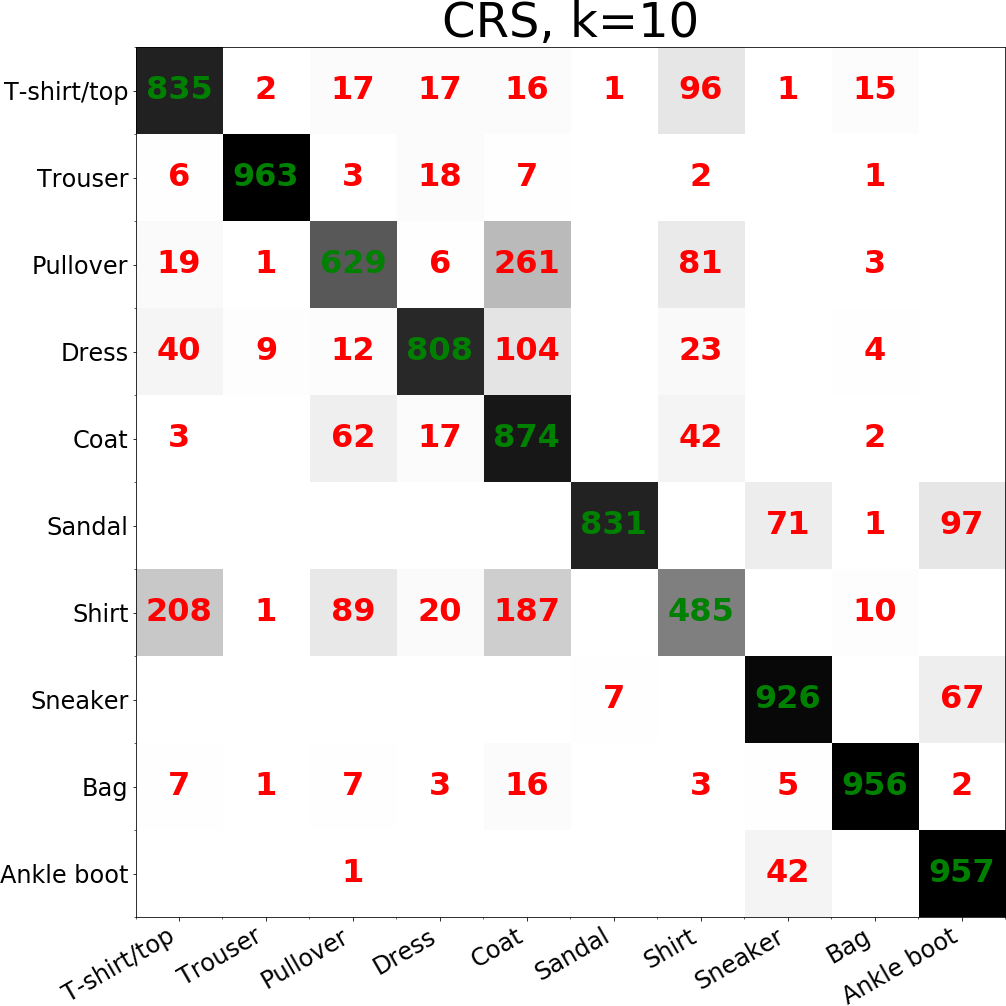}
  }%
  \subfloat[$\delta$-Medoids\label{fig:fash-delta}]{
  \includegraphics[width=0.32\textwidth]{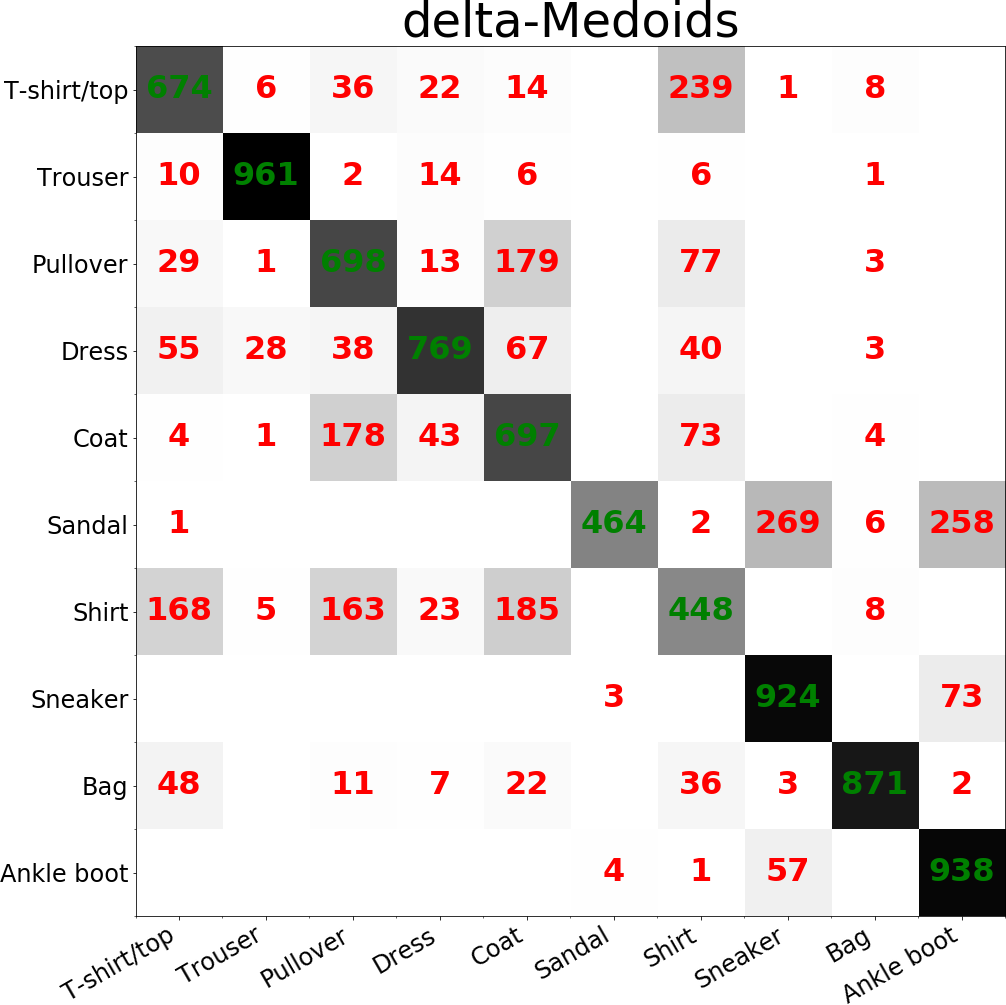}
}%
\subfloat[DS3\label{fig:fash-ds3}]{%
  \includegraphics[width=0.32\textwidth]{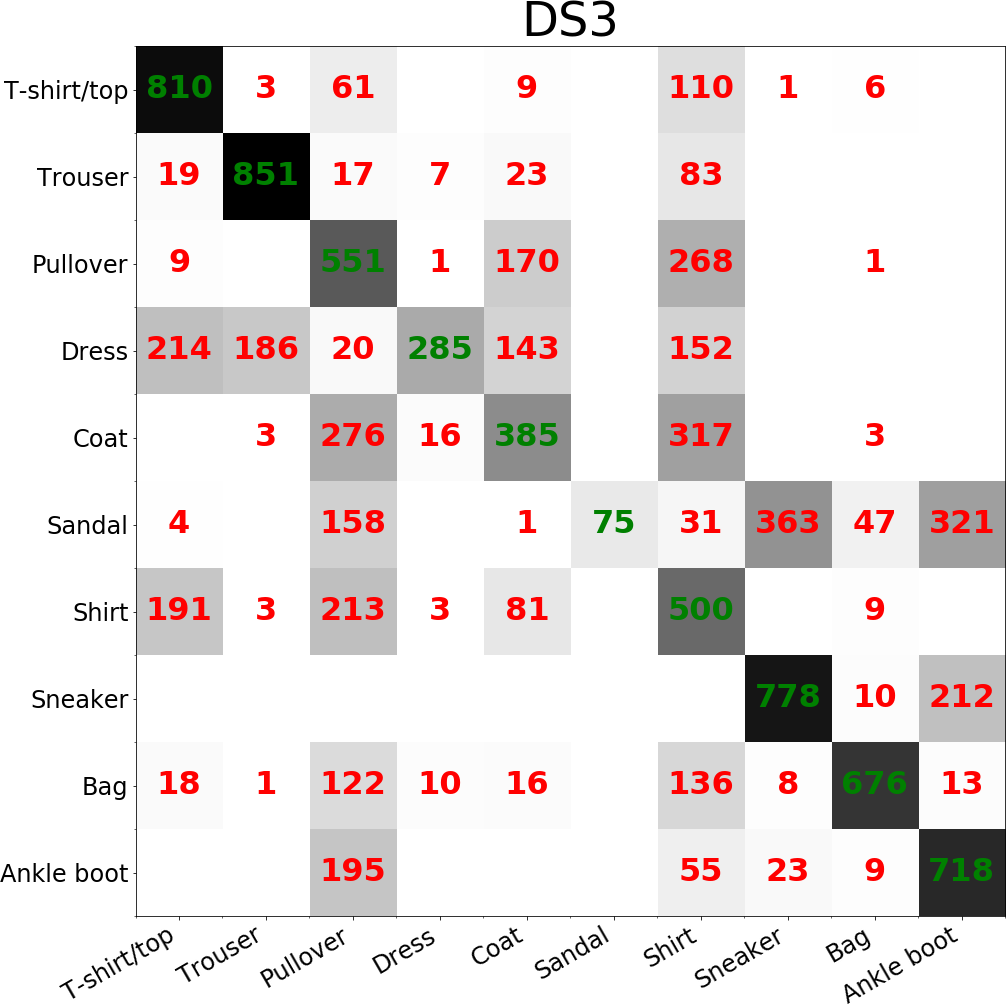}
}%
\caption{Confusion matrices for a cluster from each category in the MNIST Fashion dataset show the performance all 3 methods compared. The Sandal class was the hardest to represent for all methods. This is also quantified in Figure~\ref{fig:bar-fash}.}
\label{fig:fash}
\end{figure}

The average homogeneity of a cluster in the MNIST Fashion dataset is 0.76.
This corresponds with a slower decline of the precision and recall values as the number of representatives decreases.
In Figure~\ref{fig:bar-fash} the steep decline of representatives selected decreases only slightly decreases the precision and recall for the each cluster.
In the case of the the Dress cluster it even slightly increases from $k=10$ to $k=15$.
In Figure~\ref{fig:fash} are the confusion matrices for the methods.

\subsubsection{20Newsgroup}
\begin{figure}[h]
\centering
\subfloat[CRS-k10\label{fig:news-crs}]{%
  \includegraphics[width=0.32\textwidth]{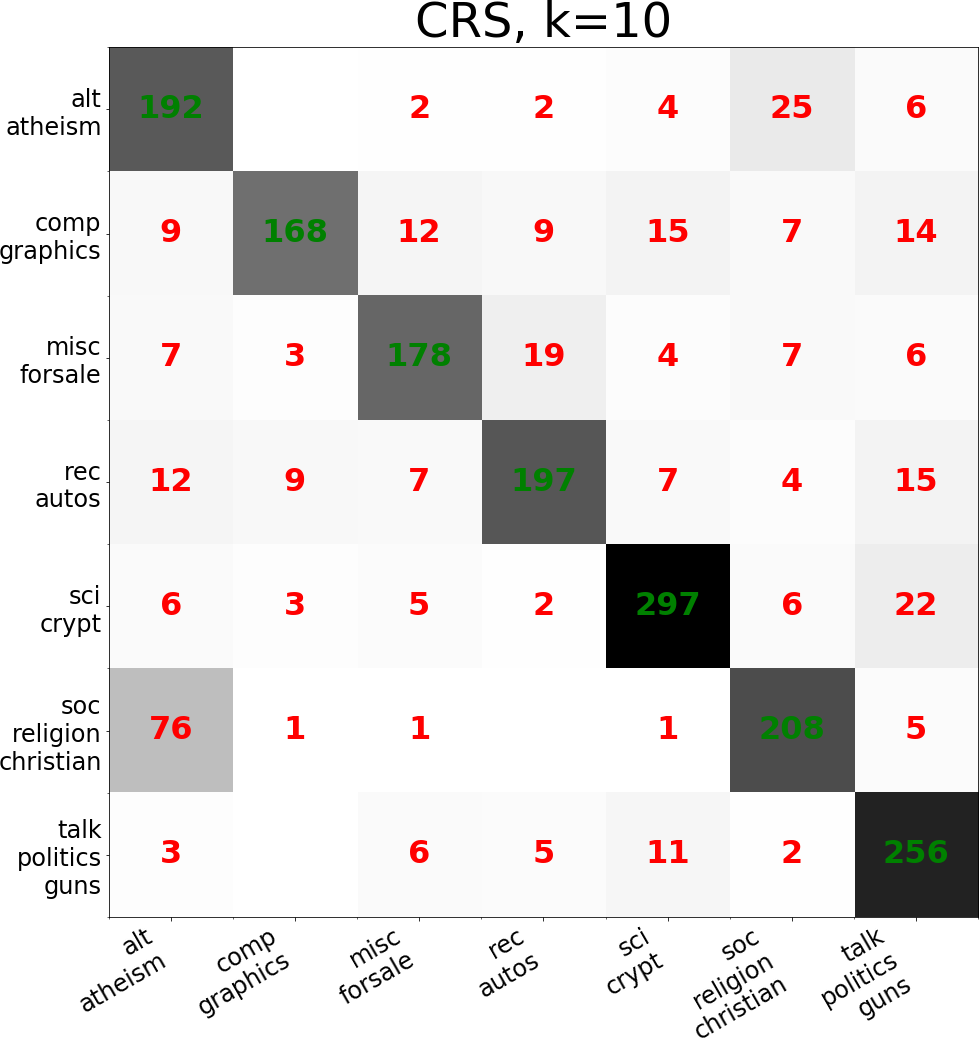}
  }%
  \subfloat[$\delta$-Medoids\label{fig:news-delta}]{
  \includegraphics[width=0.32\textwidth]{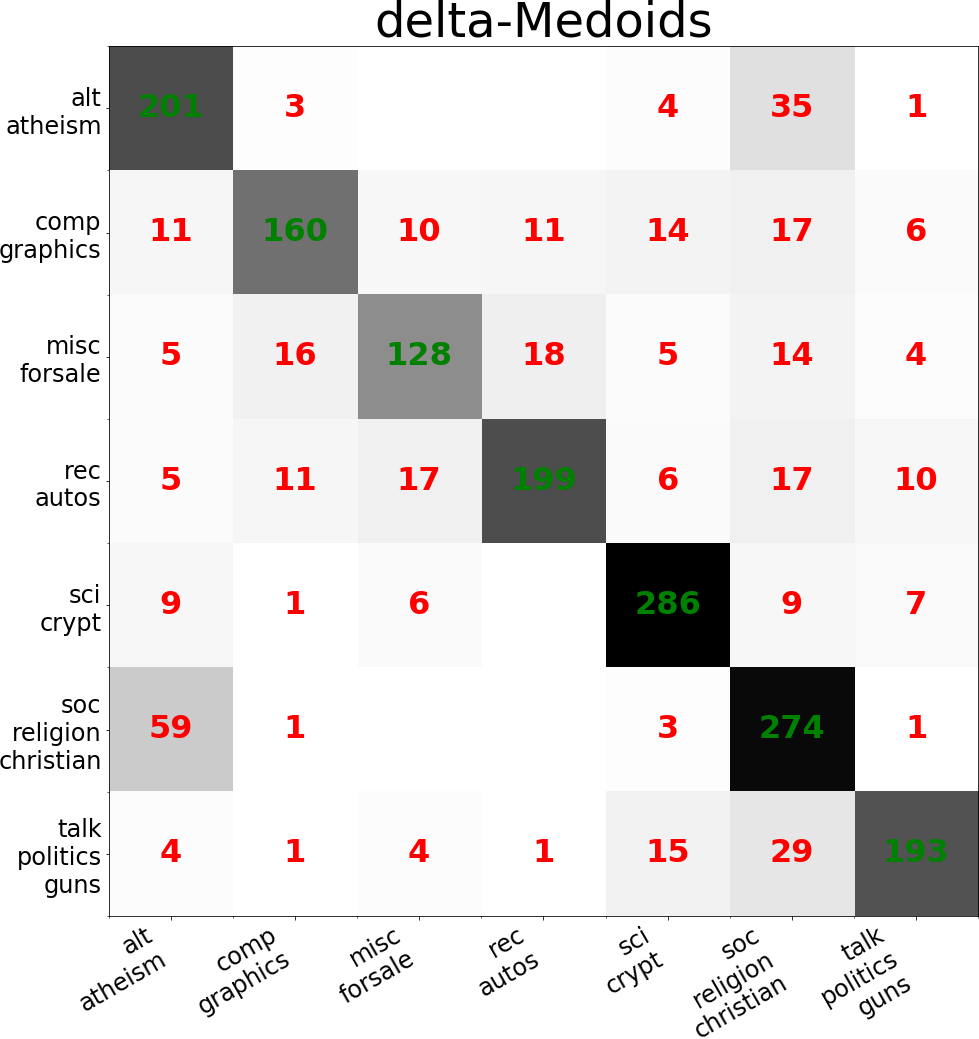}
}%
\subfloat[DS3\label{fig:news-ds3}]{%
  \includegraphics[width=0.32\textwidth]{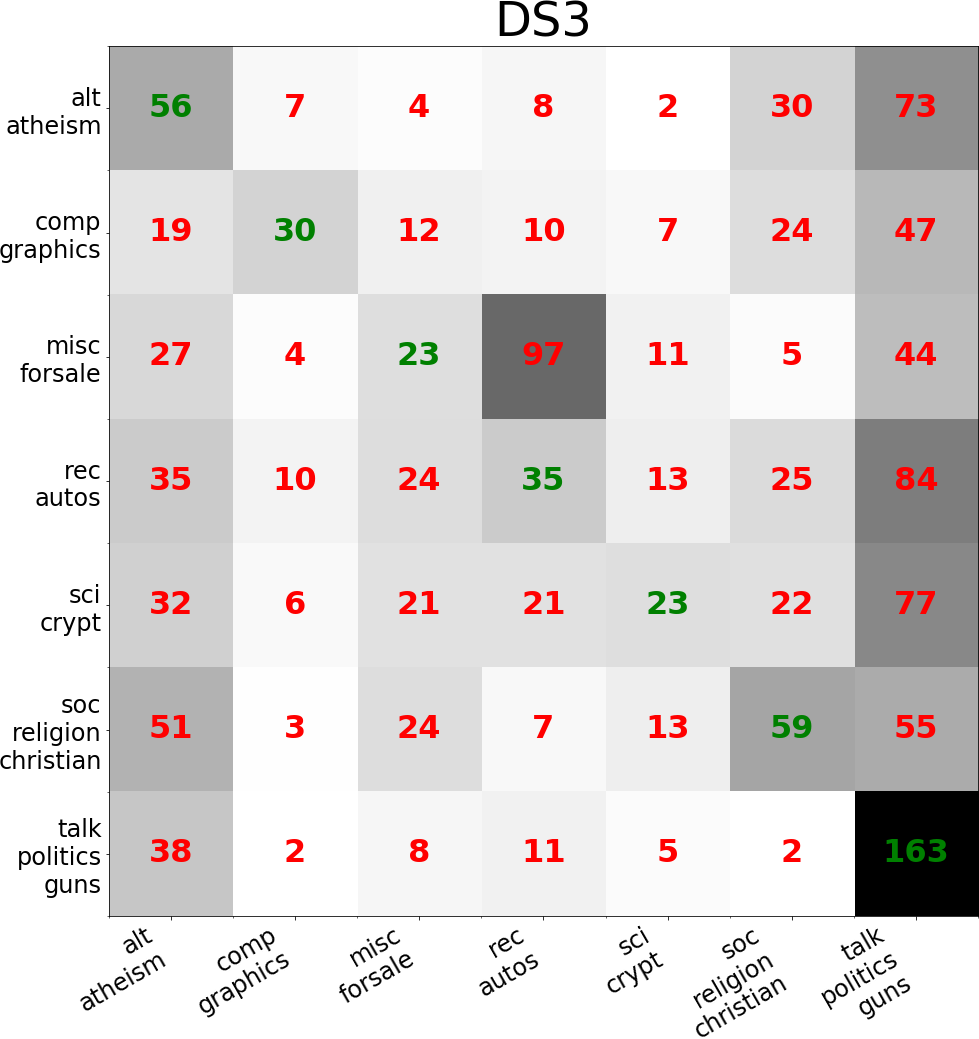}
}%
\caption{Confusion matrices for a cluster from each category in the 20Newsgroup dataset show the performance all 3 methods compared. The confusion matrix for DS3 visualizes the results from Table ~\ref{tab:all_res}}
\label{fig:news}
\end{figure}

The 20Newsgroup dataset has the lowest average homogeneity $h = 0.0858$ from all the datasets.
The samples are less similar on average, therefore the lower precision and recall values.
It reflects in the percentage of objects selected as representatives by the $\delta$-Medoids algorithm.
%Figure~\ref{fig:bar-news} shows the comparisons of algorithms for one cluster from each subgroup defined in the dataset.
% \begin{figure}
%     \centering
%     \includegraphics[width=\textwidth]{img/bar-20news.png}
%     \caption{Visualization of precision and recall of all methods in relation to percentage of cluster selected on chosen clusters for 20Newsgroup dataset. In the case of cluster \textit{rec.autos} the number the precision even increases with higher $k$.}
%     \label{fig:bar-news}
% \end{figure}
Confusion matrices for one cluster form each subgroup are in Figure~\ref{fig:news}.

\subsubsection{Network Dataset}

The results for data collected in real network further prove assumptions made in Section~\ref{sec:impact-k}.

\begin{figure}
    \centering
    \includegraphics[width=\textwidth]{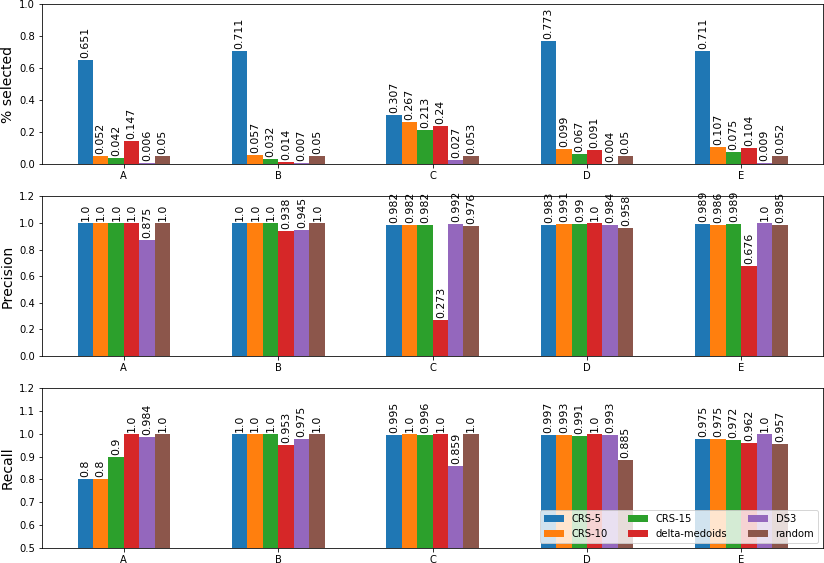}
    \caption{Visualization of precision and recall of all methods in relation to percentage of cluster selected on chosen clusters from the Network dataset.}
    \label{fig:bar-network}
\end{figure}

Figure~\ref{fig:bar-network} shows comparison of individual methods by means of precision and recall for selected large clusters as well as the impact of different values of $k$.
The depicted clusters were chosen for their sizes and different homogeneity, see Table~\ref{tab:network-sizes}.
\begin{table}[]
\begin{centering}
\begin{tabular}{|l|l|l|l|l|l|l|l|l|l|l|l|l|l|l|}
\hline
\textbf{Cluster} & \textbf{A} & \textbf{B} & \textbf{C} & \textbf{D} & \textbf{E} & \textbf{F} & \textbf{G} & \textbf{H} & \textbf{I} & \textbf{J} & \textbf{K} & \textbf{L} & \textbf{M} & \textbf{N} \\ \hline \hline
Size& 1079& 2407& 75& 2219& 346& 59& 248& 49& 52& 108& 218& 44& 42& 32\\ \hline
Homogeneity& 0.58& 0.14& 0.84& 0.64& 0.60& 0.92& 0.34& 0.84& 0.69& 0.35& 0.78& 0.35& 0.79& 1.0 \\ \hline
\end{tabular}
\end{centering}
\caption{Sizes and approximate homogeneity for each cluster from network dataset.}
\label{tab:network-sizes}
\end{table}
Increasing $k$ significantly reduces the percentage of dataset selected for its representation while still retaining high precision and recall values.
The results on the network dataset are very important because it is a non-metric dataset, where only a pair-wise similarity is defined.

\section{Conclusion}
\label{sec:conslusion}
This paper proposed a new method called CRS for building representations of clusters --- cluster prototypes which are small subsets of the original clusters.  CRS leverages nearest neighbor graphs to map structure of each cluster and to identify the most important representatives that will form the cluster prototype. Thanks to this approach, CRS can be equally applied in both metric and non-metric spaces.
The proposed method was compared to the prior art in a nearest prototype classification setup on multiple datasets from different domains. The experimental results show that the CRS method achieves superior classification quality while producing comparably compact representations of clusters.

\bibliographystyle{unsrt}  
\bibliography{references}

\begin{thebibliography}{10}

\bibitem{seo2003soft}
Sambu Seo, Mathias Bode, and Klaus Obermayer.
\newblock Soft nearest prototype classification.
\newblock {\em IEEE Transactions on Neural Networks}, 14(2):390--398, 2003.

\bibitem{schleif2005local}
F-M Schleif, Thomas Villmann, and Barbara Hammer.
\newblock Local metric adaptation for soft nearest prototype classification to
  classify proteomic data.
\newblock In {\em International Workshop on Fuzzy Logic and Applications},
  pages 290--296. Springer, 2005.

\bibitem{cervantes2007adaptive}
Alejandro Cervantes, In{\'e}s Galv{\'a}n, and Pedro Isasi.
\newblock An adaptive michigan approach pso for nearest prototype
  classification.
\newblock In {\em International Work-Conference on the Interplay Between
  Natural and Artificial Computation}, pages 287--296. Springer, 2007.

\bibitem{martino2018granular}
Alessio Martino, Alessandro Giuliani, and Antonello Rizzi.
\newblock Granular computing techniques for bioinformatics pattern recognition
  problems in non-metric spaces.
\newblock In {\em Computational Intelligence for Pattern Recognition}, pages
  53--81. Springer, 2018.

\bibitem{becker2010methods}
Glenn~C Becker.
\newblock Methods and apparatus for clustering templates in non-metric
  similarity spaces, October~12 2010.
\newblock US Patent 7,813,531.

\bibitem{kopp2018community}
Martin Kopp, Martin Grill, and Jan Kohout.
\newblock Community-based anomaly detection.
\newblock In {\em 2018 IEEE International Workshop on Information Forensics and
  Security (WIFS)}, pages 1--6. IEEE, 2018.

\bibitem{scheirer2014good}
Walter~J Scheirer, Michael~J Wilber, Michael Eckmann, and Terrance~E Boult.
\newblock Good recognition is non-metric.
\newblock {\em Pattern recognition}, 47(8):2721--2731, 2014.

\bibitem{elhamifar2015dissimilarity}
Ehsan Elhamifar, Guillermo Sapiro, and S~Shankar Sastry.
\newblock Dissimilarity-based sparse subset selection.
\newblock {\em IEEE transactions on pattern analysis and machine intelligence},
  38(11):2182--2197, 2015.

\bibitem{liebman2015representative}
Elad Liebman, Benny Chor, and Peter Stone.
\newblock Representative selection in nonmetric datasets.
\newblock {\em Applied Artificial Intelligence}, 29(8):807--838, 2015.

\bibitem{triguero2011taxonomy}
Isaac Triguero, Joaqu{\'\i}n Derrac, Salvador Garcia, and Francisco Herrera.
\newblock A taxonomy and experimental study on prototype generation for nearest
  neighbor classification.
\newblock {\em IEEE Transactions on Systems, Man, and Cybernetics, Part C
  (Applications and Reviews)}, 42(1):86--100, 2011.

\bibitem{geva1991adaptive}
Shlomo Geva and Joaquin Sitte.
\newblock Adaptive nearest neighbor pattern classification.
\newblock {\em IEEE Trans. on Neural Networks}, 2(0):2, 1991.

\bibitem{xie1993vector}
Qiaobing Xie, Charles~A Laszlo, and Rabab~K Ward.
\newblock Vector quantization technique for nonparametric classifier design.
\newblock {\em IEEE Transactions on Pattern Analysis and Machine Intelligence},
  15(12):1326--1330, 1993.

\bibitem{garcia2012prototype}
Salvador Garcia, Joaquin Derrac, Jose Cano, and Francisco Herrera.
\newblock Prototype selection for nearest neighbor classification: Taxonomy and
  empirical study.
\newblock {\em IEEE transactions on pattern analysis and machine intelligence},
  34(3):417--435, 2012.

\bibitem{wang2017representative}
Hongxing Wang, Yoshinobu Kawahara, Chaoqun Weng, and Junsong Yuan.
\newblock Representative selection with structured sparsity.
\newblock {\em Pattern Recognition}, 63:268--278, 2017.

\bibitem{zhang2018seeing}
Xingxing Zhang, Zhenfeng Zhu, Yao Zhao, Dongxia Chang, and Ji~Liu.
\newblock Seeing all from a few: l1-norm-induced discriminative prototype
  selection.
\newblock {\em IEEE transactions on neural networks and learning systems},
  30(7):1954--1966, 2018.

\bibitem{olvera2018accurate}
J~Arturo Olvera-L{\'o}pez, J~Ariel Carrasco-Ochoa, and
  J~Mart{\'\i}nez-Trinidad.
\newblock Accurate and fast prototype selection based on the notion of relevant
  and border prototypes.
\newblock {\em Journal of Intelligent \& Fuzzy Systems}, 34(5):2923--2934,
  2018.

\bibitem{dong2011efficient}
Wei Dong, Charikar Moses, and Kai Li.
\newblock Efficient k-nearest neighbor graph construction for generic
  similarity measures.
\newblock In {\em Proceedings of the 20th international conference on World
  wide web}, pages 577--586, 2011.

\bibitem{xiao2017/online}
Han Xiao, Kashif Rasul, and Roland Vollgraf.
\newblock Fashion-mnist: a novel image dataset for benchmarking machine
  learning algorithms, 2017.

\bibitem{demvsar2006statistical}
Janez Dem{\v{s}}ar.
\newblock Statistical comparisons of classifiers over multiple data sets.
\newblock {\em The Journal of Machine Learning Research}, 7(Jan):1--30, 2006.

\bibitem{shaffer1995multiple}
Juliet~Popper Shaffer.
\newblock Multiple hypothesis testing.
\newblock {\em Annual review of psychology}, 46(1):561--584, 1995.

\end{thebibliography}

\end{document}